\documentclass[11pt, a4paper]{google}

\usepackage{float}
\usepackage{makecell}        
\usepackage{fontawesome5}  
\usepackage{xspace}
\usepackage{multicol}
\usepackage{multirow}
\usepackage{longtable}
\usepackage{subcaption}
\usepackage{import}
\usepackage[usestackEOL]{stackengine}
\usepackage{framed}

\usepackage{algorithm}
\usepackage{mathtools}
\usepackage{algpseudocode}

\definecolor{headerbg}{RGB}{25, 25, 25}       
\definecolor{gridgray}{gray}{0.85}           
\definecolor{goodgreen}{RGB}{28, 142, 64}    
\definecolor{badred}{RGB}{237, 34, 39}       
\definecolor{cclnotreachedcolor}{RGB}{220, 53, 69} 

\definecolor{TargetGreen}{RGB}{46, 139, 87}

\definecolor{TargetGreen}{RGB}{46, 139, 87}

\newcolumntype{C}{>{\centering\arraybackslash}X}

\usepackage[authoryear, sort&compress, round]{natbib}
\title{AlignMamba-2: Enhancing Multimodal Fusion and Sentiment Analysis with Modality-Aware Mamba}

\author{Yan Li$^{1}$, Yifei Xing$^{1}$, Xiangyuan Lan$^{1}$, Xin Li$^{1}$, Haifeng Chen$^{2}$, Dongmei Jiang$^{3,1*}$ \\
$^{1}$ Pengcheng Laboratory, Shenzhen, China \\
$^{2}$ Shaanxi University of Science \& Technology, Xi'an, China \\
$^{3}$ School of Computer Science, Northwestern Polytechnical University, Xi'an, China
}

\let\cite\citep
\renewcommand{\today}{}

\begin{document}

\begin{abstract}
In the era of large-scale pre-trained models, effectively adapting general knowledge to specific affective computing tasks remains a challenge, particularly regarding computational efficiency and multimodal heterogeneity. While Transformer-based methods have excelled at modeling inter-modal dependencies, their quadratic computational complexity limits their use with long-sequence data. Mamba-based models have emerged as a computationally efficient alternative; however, their inherent sequential scanning mechanism struggles to capture the global, non-sequential relationships that are crucial for effective cross-modal alignment. To address these limitations, we propose \textbf{AlignMamba-2}, an effective and efficient framework for multimodal fusion and sentiment analysis. Our approach introduces a dual alignment strategy that regularizes the model using both Optimal Transport distance and Maximum Mean Discrepancy, promoting geometric and statistical consistency between modalities without incurring any inference-time overhead. More importantly, we design a Modality-Aware Mamba layer, which employs a Mixture-of-Experts architecture with modality-specific and modality-shared experts to explicitly handle data heterogeneity during the fusion process. Extensive experiments on four challenging benchmarks, including dynamic time-series (on the CMU-MOSI and CMU-MOSEI datasets) and static image-related tasks (on the NYU-Depth V2 and MVSA-Single datasets), demonstrate that AlignMamba-2 establishes a new state-of-the-art in both effectiveness and efficiency across diverse pattern recognition tasks, ranging from dynamic time-series analysis to static image-text classification.
\end{abstract}

\maketitle

\section{Introduction}
In the field of pattern recognition, recognizing complex concepts often requires integrating information from heterogeneous sources such as audio, vision, and language~\cite{li2024aves,xiao2025towards,xu2025unified}. A central problem in designing robust recognition systems lies in bridging the inherent \textit{heterogeneity gap} among these modalities, where each is characterized by distinct statistical distributions and structural patterns~\cite{xiao2024oneref,xu2026generic}. Effectively aligning and fusing these disparate data streams to generate a cohesive and comprehensive representation remains a significant and open research problem.

In the era of large-scale pre-trained models, Transformer-based architectures have established themselves as the foundation of modern AI, powering state-of-the-art Large Language Models (LLMs) and Vision Language Models (VLMs). In the context of multimodal fusion, existing methods typically leverage these powerful backbones to model complex inter-modal dependencies. These methods can be broadly categorized into two paradigms: single-stream approaches~\cite{li2019visualbert,kim2021vilt,liu2023visual}, which concatenate unimodal features and process them through a shared Transformer encoder, and multi-stream approaches~\cite{tsai2019multimodal,zheng2022multi,huang2025multimodal}, which employ dedicated encoders for each modality followed by cross-attention mechanisms for interaction. However, adapting these massive general-purpose models to affective computing tasks faces a critical bottleneck. They are fundamentally constrained by the quadratic computational complexity of the self-attention mechanism~\cite{vaswani2017attention,li2025appearance}. This limitation severely hinders their efficiency in finetuning and deployment, particularly for tasks involving long sequences or limited resources, which acts as a barrier to realizing the full potential of large-scale affective computing.

\begin{figure}[htbp]
\centering
\includegraphics[width=0.7\linewidth]{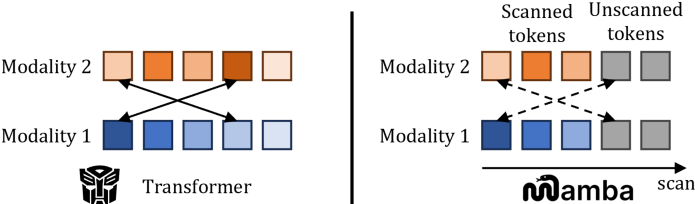}
\caption{The cross-attention mechanism in Transformer (left) and the scanning mechanism in Mamba (right).}
\label{fig:motivation}
\end{figure}

The recent introduction of State Space Models (SSMs)~\cite{gu2021combining}, particularly the Mamba architecture~\cite{gu2023mamba}, offers a promising path for the next generation of efficient large-scale models. Mamba achieves linear computational complexity while maintaining strong performance in modeling long-range dependencies. This makes it an ideal candidate to address the efficiency challenges inherent in the current large-scale model era, effectively replacing the computationally expensive Transformer backbone. This breakthrough has sparked considerable interest in adapting Mamba for multimodal fusion and sentiment analysis tasks, with approaches ranging from direct feature concatenation~\cite{qiao2024vl,zhao2025cobra} to multi-stream architectures~\cite{he2024pan,dong2024fusion,gan2025multi}. However, a direct application of Mamba to multimodal tasks reveals a critical limitation. As illustrated in Figure~\ref{fig:motivation}, Mamba's core strength, its efficient sequential scanning mechanism, becomes a fundamental weakness when modeling cross-modal relationships. The sequential scan struggles to capture the global, non-sequential dependencies between a token being processed and all tokens from other modalities, especially those that have not yet been scanned~\cite{li2025alignmamba}. This issue can lead to incomplete cross-modal information exchange and suboptimal alignment, thereby compromising the quality of the final fused representation. For example, concurrent Mamba-based multimodal methods, such as VL-Mamba~\cite{qiao2024vl} and Fusion-Mamba~\cite{dong2024fusion}, predominantly rely on direct feature concatenation or simple multi-stream interaction. While effective for general sequence modeling, these approaches lack explicit mechanisms to align heterogeneous distributions before fusion and treat the Mamba backbone as a modality-agnostic processor. Consequently, they struggle to capture the fine-grained, non-sequential cross-modal dependencies that are critical for multimodal learning and sentiment analysis.

To address these challenges, we propose a new framework, \textbf{AlignMamba-2}, which provides a holistic solution by enhancing Mamba for multimodal fusion at two crucial stages: principled alignment \textit{before} fusion and modality-aware processing \textit{during} fusion. First, we introduce a dual alignment strategy that serves as a powerful regularization term. This strategy employs two complementary distribution metrics: \textbf{Maximum Mean Discrepancy (MMD)} and \textbf{Optimal Transport (OT) distance}. MMD enforces consistency by matching the high-order statistical moments of the feature distributions, ensuring they share similar global properties. In parallel, OT distance evaluates the dissimilarity from a geometric perspective, minimizing the cost required to transform one distribution into another, thus encouraging a fine-grained alignment.
Crucially, this dual alignment strategy directly complements the inherent characteristics of Mamba's selective scan mechanism. Unlike Transformers, which can implicitly learn alignment via dense global attention maps, Mamba's sequential nature limits its ability to capture non-local, cross-modal correspondences spontaneously. By enforcing specific geometric (via OT) and statistical (via MMD) consistency prior to fusion, our strategy regularizes the latent space. This explicitly compensates for the lack of a global receptive field in the scanning process, ensuring that the computationally efficient selective scan operates on well-aligned representations, thereby maximizing the effectiveness of the linear-time fusion.
Second, and more importantly, we introduce a novel \textbf{Modality-Aware Mamba layer}. By replacing standard projection layers with a Mixture-of-Experts (MoE) structure, composed of modality-specific and modality-shared experts, our model can process tokens differently based on their origin modality. This allows the fusion backbone to explicitly capture both unique unimodal properties and shared cross-modal patterns, leading to a more effective and nuanced fusion.

In summary, the main contributions of this work are fourfold:
\begin{itemize}
    \item We present a critical analysis of existing Mamba-based multimodal methods, identifying the inherent limitations of the sequential scanning mechanism in capturing comprehensive cross-modal relationships and the modality-agnostic nature of the fusion process.
    \item We propose a dual alignment strategy using MMD and OT distance. This approach ensures robust cross-modal alignment from both statistical and geometric perspectives and, crucially, adds no extra computational cost during the inference phase.
    \item We introduce a novel Modality-Aware Mamba layer, which integrates a Mixture-of-Experts design to explicitly model both modality-specific and modality-invariant information within the fusion backbone, enabling a more sophisticated and effective fusion process.
    \item We conduct experiments on a diverse set of multimodal fusion and sentiment analysis benchmarks, including dynamic tasks (on CMU-MOSI and CMU-MOSEI datasets) and static tasks (on NYU-Depth V2 and MVSA-Single datasets), demonstrating the superior performance and broad applicability of AlignMamba-2.
\end{itemize}

This paper is an extended version of our preliminary work presented in~\cite{li2025alignmamba}. While the foundational idea of leveraging Mamba for aligned multimodal fusion is shared, this work introduces substantial advancements and a much broader scope of analysis. Specifically, the key extensions are threefold: 
\textbf{(1)} We replace the explicit OT matrix computation from our prior work with the OT distance-based loss. This architectural refinement not only simplifies the model but also eliminates all alignment-related computational overhead during inference, enhancing the model's practicality for real-world deployment. 
\textbf{(2)} We move beyond simple pre-fusion alignment and introduce a novel Modality-Aware Mamba layer. This core architectural innovation endows the fusion backbone itself with the ability to process modality-specific and modality-invariant information, addressing a key limitation of modality-agnostic sequence models.
\textbf{(3)} We significantly expand the empirical validation of our framework. In addition to the original multimodal sentiment analysis tasks, we now include comprehensive evaluations on the out-of-distribution setting and static multimodal tasks, namely RGB-D scene recognition and image-text classification, thereby demonstrating the versatility and generalizability of our approach across diverse data types and applications.
\section{Related Work}
In this section, we review prior work from four perspectives relevant to our proposed method: Transformer-based multimodal fusion, Mamba-based multimodal fusion, the Mixture-of-Experts paradigm, and techniques for multimodal representation alignment.

\subsection{Transformer-based Multimodal Fusion}
The Transformer architecture~\cite{vaswani2017attention}, with its powerful self-attention mechanism, has long dominated the field of multimodal representation learning. Existing approaches can be broadly classified into two main streams. \textbf{Single-stream} models, such as VisualBERT~\cite{li2019visualbert} and ViLT~\cite{kim2021vilt}, concatenate feature sequences from different modalities into a single sequence and process it through a unified Transformer encoder. This paradigm facilitates late fusion by allowing tokens from all modalities to interact within the same attention space~\cite{zhang2023cross}. In contrast, \textbf{multi-stream} models like MulT~\cite{tsai2019multimodal} and MM-PEAR-CoT~\cite{li2025multimodal} maintain separate backbones for each modality and employ cross-attention layers~\cite{huang2025multimodal} to enable explicit, pairwise information exchange between them. These methods have demonstrated strong performance by capturing intricate cross-modal dependencies. However, their reliance on the self-attention mechanism, which has a computational complexity quadratic to the input sequence length, creates a fundamental efficiency bottleneck. This makes them prohibitively expensive for long-sequence tasks and resource-constrained environments.

In contrast, AlignMamba-2 directly addresses this fundamental efficiency bottleneck by adopting a linear-time Mamba backbone, while simultaneously introducing mechanisms to overcome Mamba's inherent limitations in cross-modal modeling.

\subsection{Mamba-based Multimodal Fusion}
Inspired by the success of Mamba in natural language processing~\cite{gu2023mamba}, a growing body of work has explored its application to multimodal tasks. These methods often adopt straightforward strategies, such as concatenating multimodal features for a single Mamba backbone~\cite{qiao2024vl,zhao2025cobra} or using multi-stream designs~\cite{he2024pan,dong2024fusion,gan2025multi}. However, they face two significant challenges. First, as discussed, the inherent sequential scanning mechanism of Mamba makes it difficult to model the complex, non-sequential relationships between tokens across different modalities. Our previous work, AlignMamba-1~\cite{li2025alignmamba}, partially addressed this by introducing an explicit pre-fusion alignment module. Second, existing approaches typically treat the Mamba block as a generic, modality-agnostic sequence processor. This overlooks the fact that different modalities possess unique intrinsic properties that may require specialized processing during the fusion stage~\cite{bao2022vlmo}.

Our work, AlignMamba-2, provides a more comprehensive solution. It not only employs a principled alignment strategy to address the scanning limitation but also redesigns the fusion core itself with a Modality-Aware Mamba layer to handle data heterogeneity, a capability absent in prior Mamba-based models.

\subsection{Mixture-of-Experts}
The Mixture-of-Experts paradigm is a powerful technique for increasing a model's capacity without a proportional increase in computational cost. An MoE layer consists of multiple "expert" subnetworks and a "gating" network that sparsely activates a subset of these experts for each input token. This approach has been successfully employed to scale up large language models to trillions of parameters, such as in GShard~\cite{lepikhin2020gshard}, Switch Transformers~\cite{fedus2022switch}, and DeepSeek-MoE~\cite{dai2024deepseekmoe}. In these applications, the primary goal is to expand model knowledge and capability while keeping inference costs manageable through learnable routing.

Our work introduces this powerful architecture into the Mamba framework for a novel purpose. We design a Modality-Aware Mamba layer that features a deterministic MoE structure composed of both \textbf{modality-specific experts} and a \textbf{modality-shared expert}. This design enables the model to simultaneously learn modality-specific characteristics and general cross-modal patterns, leading to a more comprehensive and effective multimodal fusion representation.

\subsection{Multimodal Representation Alignment}
Bridging the heterogeneity gap between modalities necessitates effective representation alignment, which can be broadly categorized into implicit and explicit approaches. \textbf{Implicit alignment} methods learn a shared embedding space by optimizing a specific objective function, without computing an explicit token-to-token correspondence matrix~\cite{dai2024multimodal}. This includes contrastive learning methods~\cite{mai2022hybrid,lin2023multi}, which pull representations of corresponding (positive) multimodal pairs together~\cite{li2023towards}. Another prominent direction is distribution matching. Techniques like Deep Canonical Correlation Analysis (DCCA) aim to maximize the correlation between the latent representations of different modalities~\cite{sun2020learning}, while Maximum Mean Discrepancy directly minimizes the distance between feature distributions in a Reproducing Kernel Hilbert Space (RKHS). These methods enforce global statistical consistency between modalities. \textbf{Explicit alignment} methods, on the other hand, focus on finding specific correspondences. A notable example is Optimal Transport~\cite{villani2021topics}, which computes a transport plan (an explicit matrix) that details the most efficient way to map tokens from one modality to another~\cite{li2025alignmamba,cao2022otkge}. While this approach provides strong, interpretable alignment, the computation of the transport matrix can introduce significant computational overhead, particularly for long sequences.

In contrast, we integrate a dual alignment strategy, using MMD and OT distance as regularization losses, directly with the Mamba architecture. This approach not only provides a robustly aligned foundation from both statistical and geometric perspectives but, crucially, does so entirely during the training phase. By formulating alignment as a training objective rather than a computational step at inference, our method ensures that Mamba can effectively process aligned features without incurring any extra computational cost once the model is deployed.
\section{Methodology}
In this section, we first provide a brief overview of the Mamba architecture as a preliminary. We then present the overall framework of our proposed AlignMamba-2. Subsequently, we detail its core components: the unimodal encoding and alignment strategy, and the novel Modality-Aware Mamba layer for fusion. Finally, we outline the training objective and provide a summary of the algorithm.

\subsection{Preliminaries: The Mamba Architecture}
The Mamba architecture~\cite{gu2023mamba} is built upon a structured State Space Model (SSM)~\cite{gu2021combining}, which maps a 1-D input sequence $u(t) \in \mathbb{R}$ to an output $y(t) \in \mathbb{R}$ through a latent state $h(t) \in \mathbb{R}^N$. The continuous-time formulation is given by:
\begin{equation}
\begin{aligned}
    h'(t) &= \mathbf{A}h(t) + \mathbf{B}u(t), \\
    y(t) &= \mathbf{C}h(t),
\end{aligned}
\end{equation}
where $\mathbf{A} \in \mathbb{R}^{N \times N}$ is the state transition matrix, and $\mathbf{B}, \mathbf{C} \in \mathbb{R}^{N \times 1}$ are projection matrices. For use in deep learning models, these continuous parameters are discretized using a timestep $\Delta$. A common discretization method is the zero-order hold (ZOH), which yields discrete parameters $\overline{\mathbf{A}}$ and $\overline{\mathbf{B}}$. Mamba introduces a key innovation: the \textit{selective scan mechanism}, which makes the parameters $\mathbf{B}$, $\mathbf{C}$, and $\Delta$ input-dependent functions of the input sequence $x \in \mathbb{R}^{T \times d}$. This allows the model to selectively focus on or ignore information at each timestep. The vanilla Mamba layer packages this SSM into a cohesive unit with input ($\text{Linear}_{\text{in}}$) and output ($\text{Linear}_{\text{out}}$) projections, a causal convolution, and gating mechanisms, making it a powerful and efficient replacement for the Transformer's attention layer.

\subsection{Overall Framework}
\begin{figure*}[htbp]
\centering
\includegraphics[width=\linewidth]{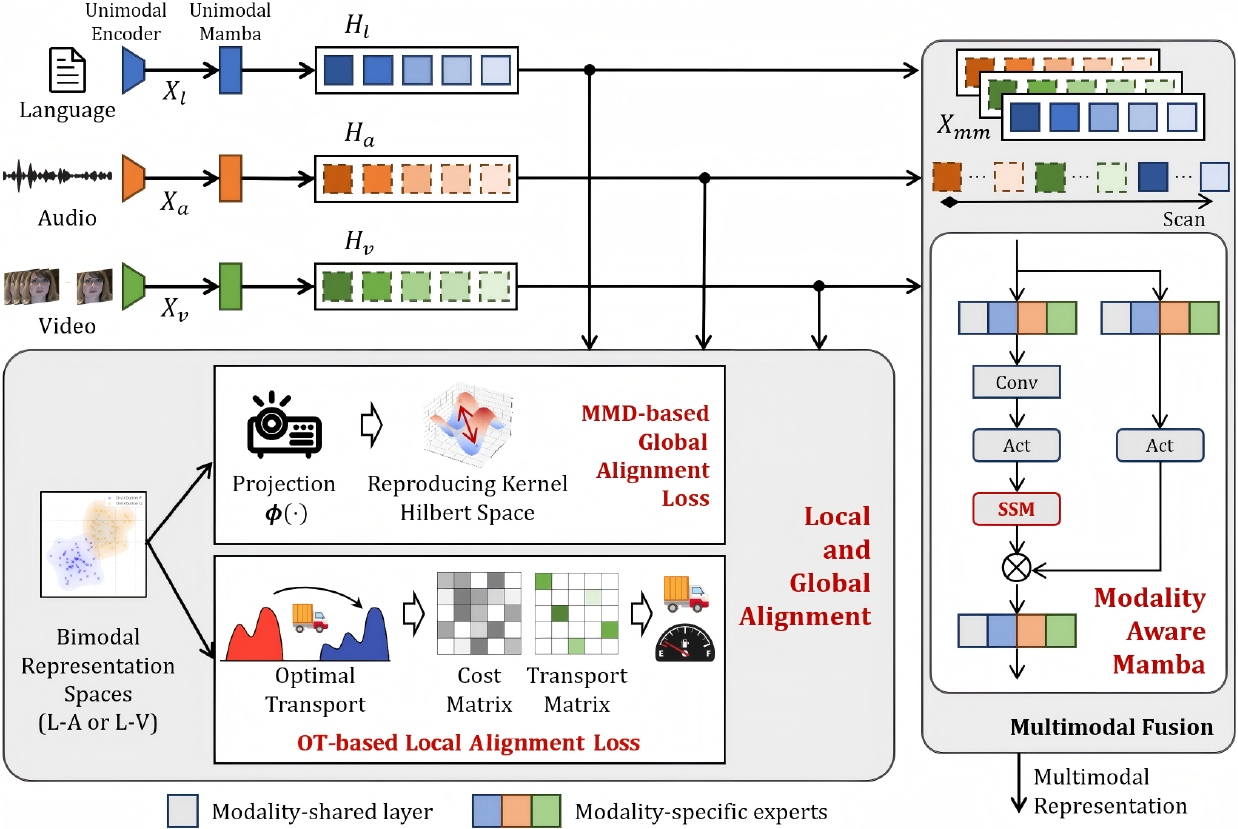}
\caption{The architecture of AlignMamba-2. The framework consists of three main stages: unimodal encoding, training-time dual alignment regularization (using OT and MMD), and a novel modality-aware Mamba fusion module based on a Mixture-of-Experts design.}
\label{fig:framework}
\end{figure*}

As illustrated in Figure~\ref{fig:framework}, our proposed AlignMamba-2 framework is designed to perform robust and efficient multimodal fusion. Using a trimodal scenario with audio, video, and language for illustration, the process begins by feeding raw data from each modality into its respective unimodal encoder to extract high-level feature sequences. These sequences are then processed by unimodal vanilla Mamba layers to model intra-modal contextual dependencies. Crucially, at the output of these unimodal Mamba layers, we apply our dual alignment strategy, which imposes both local geometric and global statistical constraints on the representations, prompting the alignment before fusion. Subsequently, the aligned feature sequences from all modalities are concatenated and fed into a stack of our Modality-Aware Mamba layers. These layers are specifically designed to handle heterogeneous data by jointly learning modality-specific and modality-invariant patterns, leading to a comprehensive fused representation. Finally, this representation is passed to a prediction head for the downstream task, such as sentiment analysis or scene classification.

\subsection{Unimodal Encoding and Alignment}
This stage aims to generate contextually-aware and well-aligned unimodal representations, laying a solid foundation for the subsequent fusion process.

\textbf{Unimodal Encoding.}
Let the input feature sequences for audio, video, and language be denoted as $X_m \in \mathbb{R}^{T_m \times d_{in}}$ for $m \in \{A, V, L\}$. Each sequence is first processed by a shallow unimodal Mamba block to capture its internal temporal dependencies:
\begin{equation}
    H_m = \text{Mamba}_{\text{m}}(X_m) \in \mathbb{R}^{T_m \times d}.
\end{equation}
The resulting representations $\{H_A, H_V, H_L\}$ are then subjected to our dual alignment constraints. We use the language modality ($H_L$) as the anchor, aligning audio ($H_A$) and video ($H_V$) representations to it.

\textbf{Local Alignment via OT Distance.}
To capture fine-grained, token-level correspondences, we use Optimal Transport (OT) distance~\cite{villani2021topics} as a local alignment loss. Given two empirical distributions over token representations from modalities $m$ and $n$, defined as $\mu = \frac{1}{T_m}\sum_{i=1}^{T_m}\delta_{h_{m,i}}$ and $\nu = \frac{1}{T_n}\sum_{j=1}^{T_n}\delta_{h_{n,j}}$, the OT distance is formulated as:
\begin{equation}
\label{eq:ot_problem}
W(\mu, \nu) = \min_{\mathbf{P} \in \Pi(\mu, \nu)} \sum_{i=1}^{T_m}\sum_{j=1}^{T_n} P_{ij} C_{ij},
\end{equation}
where $\mathbf{P}$ is the transport plan in the set of all valid plans $\Pi(\mu, \nu)$, and $\mathbf{C}$ is the ground cost matrix. We define the cost $C_{ij}$ as the cosine distance between tokens:
\begin{equation}
C_{ij} = 1 - \frac{h_{m,i} \cdot h_{n,j}}{\|h_{m,i}\| \|h_{n,j}\|}.
\end{equation}
Solving Eq.~\ref{eq:ot_problem} exactly is computationally expensive. Therefore, we use the Sinkhorn algorithm~\cite{cuturi2013sinkhorn}, which adds an entropic regularization term to make the problem efficiently solvable:
\begin{equation}
\mathcal{L}_{\text{OT}}(H_m, H_n) = \min_{\mathbf{P} \in \Pi(\mu, \nu)} \left( \langle \mathbf{P}, \mathbf{C} \rangle_F - \epsilon H(\mathbf{P}) \right),
\label{eq:epsilon}
\end{equation}
where $\epsilon > 0$ is the regularization strength and $H(\mathbf{P}) = -\sum_{i,j}P_{ij}\log(P_{ij})$ is the entropy of the plan. This loss penalizes misalignment by encouraging tokens with similar semantics (low cost) to be matched.

\textbf{Global Alignment via MMD.}
To prompt consistency at a global, statistical level, we use the Maximum Mean Discrepancy (MMD)~\cite{gretton2012kernel}. MMD measures the distance between two distributions as the squared norm of the difference between their mean embeddings in a Reproducing Kernel Hilbert Space (RKHS), $\mathcal{H}$. The empirical estimate of the squared MMD between representations $H_m$ and $H_n$ is:

\begin{equation}
\begin{multlined}
    \mathcal{L}_{\text{MMD}}(H_m, H_n) = \left\| \frac{1}{T_m}\sum_{i=1}^{T_m}\phi(h_{m,i}) - \frac{1}{T_n}\sum_{j=1}^{T_n}\phi(h_{n,j}) \right\|_{\mathcal{H}}^2 \\
    = \frac{1}{T_m^2}\sum_{i,i'}k(h_{m,i}, h_{m,i'}) - \frac{2}{T_m T_n}\sum_{i,j}k(h_{m,i}, h_{n,j}) + \frac{1}{T_n^2}\sum_{j,j'}k(h_{n,j}, h_{n,j'}),
\end{multlined}
\end{equation}
where $\phi(\cdot)$ is the feature map to the RKHS and $k(\cdot, \cdot)$ is the associated kernel function. Here, we use the common Gaussian kernel:
\begin{equation}
    k(x, y) = \exp(-\|x-y\|^2 / (2\sigma^2)).
\end{equation}
This loss pulls the high-order statistical moments of the distributions closer, ensuring they are globally indistinguishable.

Our full alignment loss
\begin{equation}
    \mathcal{L}_{\text{align}} = \lambda_{\text{OT}}\mathcal{L}_{\text{OT}} + \lambda_{\text{MMD}}\mathcal{L}_{\text{MMD}}
\end{equation}
is applied to both the video-language and audio-language pairs. The two losses are complementary: OT enforces local geometric similarity, while MMD ensures global statistical consistency.

\subsection{Modality-Aware Mamba for Fusion}
\begin{figure}[htbp]
\centering
\includegraphics[width=0.5\linewidth]{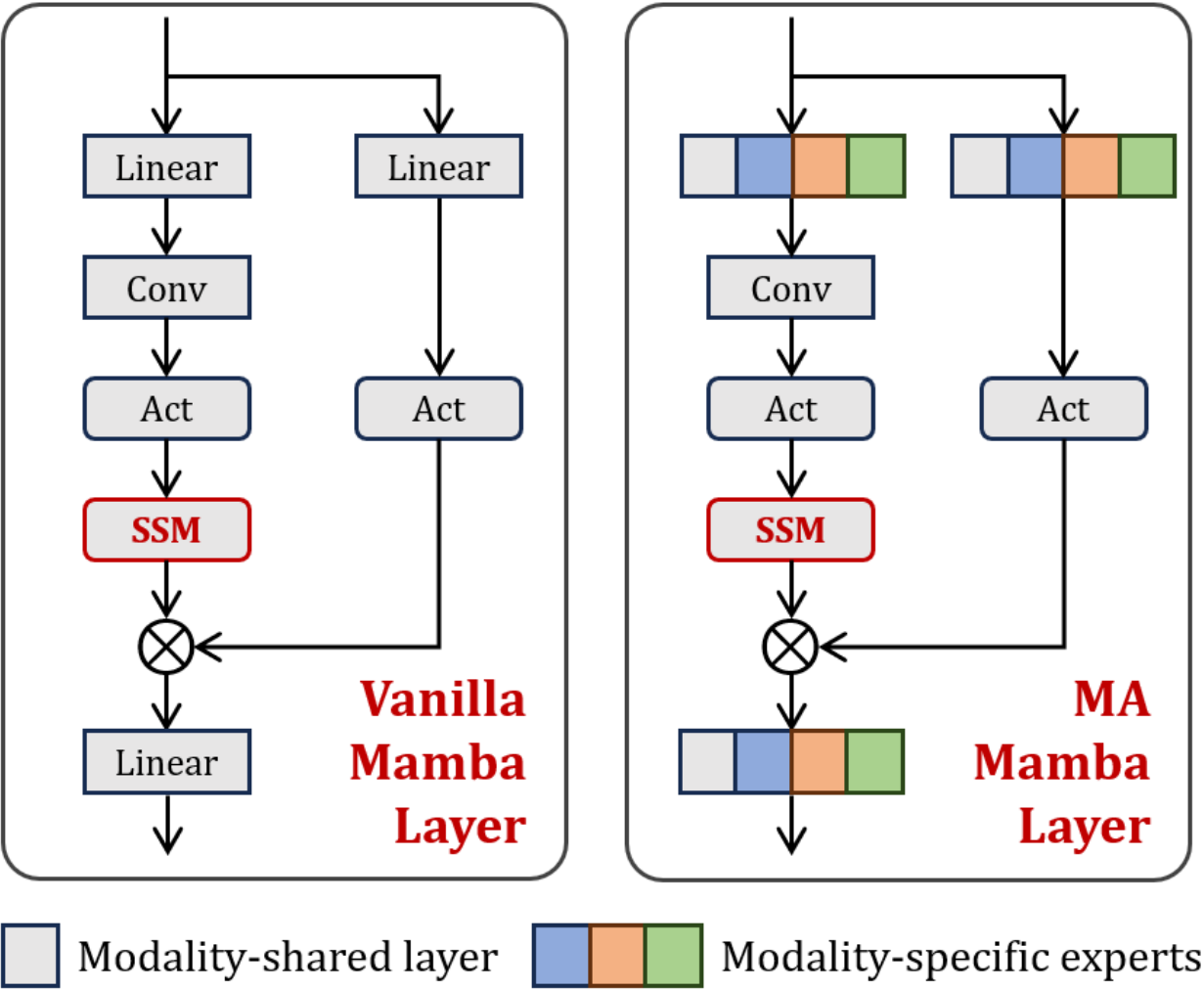}
\caption{Vanilla Mamba layer (left) and Modality-Aware Mamba layer (right).}
\label{fig:ma_mamba}
\end{figure}

A key limitation of vanilla Mamba in multimodal settings is its modality-agnostic nature. The shared projection and SSM layers process all tokens identically, regardless of their origin modality. To overcome this, we introduce the Modality-Aware Mamba layer, as depicted in Figure~\ref{fig:ma_mamba}. Our motivation is to equip the Mamba block with the ability to learn both modality-specific features and shared cross-modal patterns simultaneously.

We achieve this by replacing the standard linear input and output projection layers ($\text{Linear}_{\text{in}}, \text{Linear}_{\text{out}}$) with a Mixture-of-Experts (MoE) structure. For a concatenated input sequence $H_{\text{concat}} = [H_A; H_V; H_L]$, a token $h_i$ belonging to modality $m$ is routed deterministically based on its known origin.

Specifically, the MoE-based input projection consists of a set of modality-specific experts $\{E_{\text{in}, m}\}_{m \in \{A,V,L\}}$ and one shared expert $E_{\text{in}, \text{shared}}$. All experts have an identical architecture. The projection for a token $h_i$ from modality $m$ is:
\begin{equation}
    \text{MoE}_{\text{in}}(h_i) = E_{\text{in}, m}(h_i) + E_{\text{in}, \text{shared}}(h_i).
\end{equation}
Similarly, the output projection is also an MoE layer with experts $\{E_{\text{out}, m}\}$ and $E_{\text{out}, \text{shared}}$:
\begin{equation}
    \text{MoE}_{\text{out}}(h_i) = E_{\text{out}, m}(h_i) + E_{\text{out}, \text{shared}}(h_i).
\end{equation}
Crucially, the intermediate components, e.g., the 1D convolution, SiLU activation, and the selective SSM core, remain shared across all modalities. This design strikes a balance: the experts capture modality-specific transformations, while the shared core learns the universal dynamics of sequence modeling across all modalities. This enables a far more nuanced and effective fusion process than a standard Mamba block.

\subsection{Training Objective and Algorithm}
The entire AlignMamba-2 model is trained end-to-end. The final training objective is a sum of the task-specific loss and our alignment loss:
\begin{equation}
    \mathcal{L}_{\text{total}} = \mathcal{L}_{\text{task}} + \mathcal{L}_{\text{align}},
\end{equation}
where $\mathcal{L}_{\text{task}}$ is the downstream loss (e.g., cross-entropy for classification, Mean Absolute Error for regression). The complete training procedure is summarized in Algorithm~\ref{alg:alignmamba2}.

\begin{algorithm}[htbp]
\caption{Training Process for AlignMamba-2}
\label{alg:alignmamba2}
\begin{algorithmic}[1]
\State \textbf{Input:} Raw data for each modality $\{X_m^{\text{raw}}\}_{m \in \{A,V,L\}}$, labels $Y$.
\State \textbf{Parameters:} Unimodal encoders $\{E_m\}$, unimodal Mamba $\{Mamba_m\}$, Modality-Aware Mamba $MAMamba$, prediction head $C$, weights $\lambda_{\text{OT}}, \lambda_{\text{MMD}}$.
\State
\For{each modality $m \in \{A, V, L\}$}
    \State $X_m \gets E_m(X_m^{\text{raw}})$ \Comment{Extract unimodal features}
    \State $H_m \gets Mamba_m(X_m)$ \Comment{Model intra-modal context}
\EndFor
\State
\State $\mathcal{L}_{\text{OT}} \gets \mathcal{L}_{\text{OT}}(H_A,H_L) + \mathcal{L}_{\text{OT}}(H_V,H_L)$ \Comment{Compute total OT loss}
\State $\mathcal{L}_{\text{MMD}} \gets \mathcal{L}_{\text{MMD}}(H_A,H_L) + \mathcal{L}_{\text{MMD}}(H_V,H_L)$ \Comment{Compute total MMD loss}
\State
\State $H_{\text{concat}} \gets \text{Concatenate}(H_A, H_V, H_L)$ \Comment{Concatenate unimodal features}
\State $Z \gets MAMamba(H_{\text{concat}})$ \Comment{Modality-aware fusion}
\State $\hat{Y} \gets C(Z)$ \Comment{Make prediction}
\State
\State $\mathcal{L}_{\text{task}} \gets \text{CrossEntropy / MAE}(Y, \hat{Y})$ \Comment{Compute task-specific loss}
\State $\mathcal{L}_{\text{total}} \gets \mathcal{L}_{\text{task}} + \lambda_{\text{OT}}\mathcal{L}_{\text{OT}} + \lambda_{\text{MMD}}\mathcal{L}_{\text{MMD}}$
\State Perform backpropagation on $\mathcal{L}_{\text{total}}$.
\end{algorithmic}
\end{algorithm}
\section{Experiments}
In this section, we conduct a comprehensive set of experiments to validate the effectiveness and efficiency of AlignMamba-2. We first introduce the datasets and implementation details (Section~\ref{sec:datasets}). We then present our main results on both dynamic multimodal tasks (Section~\ref{sec:dynamic}) and static multimodal tasks (Section~\ref{sec:static}). Next, we provide an in-depth analysis of the model's computational efficiency (Section~\ref{sec:efficiency}), conduct ablation studies to dissect the contribution of each component (Section~\ref{sec:ablation}), and evaluate the model's cross-dataset generalization capabilities (Section~\ref{sec:generalization}). Finally, we analyze the impact of key hyperparameters (Section~\ref{sec:hyperparams}).

\subsection{Datasets and Implementation Details}
\label{sec:datasets}
\paragraph{Datasets}
To demonstrate the versatility and generalizability of our proposed AlignMamba-2, we evaluate it on four challenging multimodal fusion and sentiment analysis benchmarks spanning both dynamic (time-series) and static (image-based) settings.

\begin{itemize}
\item \textbf{CMU-MOSI (MOSI)}~\cite{zadeh2016mosi}: A widely-used benchmark for multimodal sentiment analysis. It consists of 2,199 short video clips where aligned language, visual, and acoustic streams are available.
While the original annotations are on a scale from -3 to +3, we follow the standard evaluation protocol by framing the task as a binary classification (positive vs. negative).
\item \textbf{CMU-MOSEI (MOSEI)}~\cite{zadeh2018multimodal}: A larger and more complex dataset for multimodal sentiment analysis, containing 22,856 video clips.
MOSEI presents greater challenges due to its larger vocabulary, more diverse topics, and more complex emotional expressions. Similar to MOSI, we adhere to the standard binary classification setup (positive vs. negative).
\item \textbf{NYU-Depth v2 (NYUDv2)}~\cite{silberman2012indoor}: A popular indoor scene understanding dataset that provides tightly-coupled RGB and depth (D) modalities for each image.
To ensure a fair comparison with prior works, we follow the evaluation setup from~\cite{gao2024embracing} and frame the task as a 10-class scene recognition problem. This involves using 9 of the most common scene categories and grouping all remaining categories into a single "Others" class.
\item \textbf{MVSA-Single (MVSA)}~\cite{niu2016sentiment}: A widely-used benchmark for image-text sentiment analysis. It contains over 4,800 image-text pairs sourced from Twitter, with each pair labeled as positive, negative, or neutral.
A key characteristic of this dataset is the potential for semantic conflicts between the visual and textual modalities, posing a significant challenge for multimodal fusion methods.
\end{itemize}

Following previous work~\cite{gao2024embracing,li2025alignmamba}, we adopt accuracy and F1 score as evaluation metrics on these four datasets.

\paragraph{Implementation Details}
To ensure a fair comparison, we adopt the experimental setup from previous methods~\cite{zadeh2016mosi,zadeh2018multimodal,gao2024embracing}. Specifically, for the unimodal encoders on the MOSI and MOSEI datasets, we utilize the officially provided features extracted by FACET (for visual modality) and COVAREP (for audio modality) to maintain consistency with prior work. For the NYUDv2 and MVSA datasets, we employ a pre-trained ResNet as the encoder for both the RGB and depth modalities. For the text modality, we use the Hugging Face implementation of the Mamba model.

Our model is implemented using PyTorch. We use the Adam optimizer with an initial learning rate of 1e-3 and employ a ReduceLROnPlateau learning rate scheduler. The gradient clipping parameter is set to 1.0. The number of layers for both the unimodal and modality-aware Mamba is selected from 1 to 3. To prevent overfitting, we apply a dropout rate of 0.2 and utilize an early stopping strategy.

Regarding the hyperparameter configurations for the dual alignment strategy, we specify the details as follows. For the MMD loss, we employ a Gaussian kernel, where the kernel bandwidth $\gamma$ is set to the inverse of the input feature dimension $d$ (i.e., $\gamma = 1/d$) to ensure scale invariance across different modalities. For the OT loss, we utilize the Sinkhorn algorithm implemented via the GeomLoss library. The entropic regularization strength $\epsilon$ (as denoted in Eq.~\ref{eq:epsilon}) is controlled by setting the blur radius parameter to $0.05$. Furthermore, instead of fixing a static number of iterations, we employ the library's automatic epsilon-scaling mechanism with a tensorized backend. This adaptive strategy dynamically performs sufficient Sinkhorn iterations to guarantee numerical convergence and stability during training.

\subsection{Results on MOSI and MOSEI}
\label{sec:dynamic}
We present the results of AlignMamba-2 on the dynamic multimodal fusion benchmarks, CMU-MOSI and CMU-MOSEI, in Table~\ref{tab:mosi_mosei_results}. We compare our method against a wide range of state-of-the-art approaches, which can be broadly categorized into two groups: Transformer-based methods and the more recent Mamba-based methods.

\begin{table}[t]
\centering
\caption{Results on the MOSI and MOSEI datasets.}
\label{tab:mosi_mosei_results}
\begin{tabular}{l|r|cc|cc}
\toprule
\multicolumn{1}{c|}{\multirow{2}{*}{\textbf{Method}}} & \multirow{2}{*}{\textbf{Publication}} & \multicolumn{2}{c|}{\textbf{MOSI}} & \multicolumn{2}{c}{\textbf{MOSEI}} \\
\multicolumn{1}{c|}{} & & Acc & F1 & Acc & F1 \\
\midrule
\multicolumn{6}{l}{\textit{Transformer-based Methods}} \\
\midrule
MulT \cite{tsai2019multimodal} & ACL'19 & 84.1 & 83.9 & 82.5 & 82.3 \\
ICCN \cite{sun2020learning} & AAAI'20 & 83.0 & 83.0 & 84.2 & 84.2 \\
MISA \cite{hazarika2020misa} & MM'20 & 83.4 & 83.6 & 85.5 & 85.3 \\
Self-MM \cite{yu2021learning} & AAAI'21 & 84.8 & 84.9 & 85.0 & 84.9 \\
MMIM \cite{han2021improving} & EMNLP'21 & 85.1 & 85.0 & 85.1 & 85.0 \\
TokenFusion \cite{wang2022multimodal} & CVPR'22 & 84.8 & 84.7 & 85.0 & 84.9 \\
HyCon \cite{mai2022hybrid} & TAFFC'22 & 85.2 & 85.1 & 85.4 & 85.6 \\
ConFEDE \cite{yang2023confede} & ACL'23 & 85.5 & 85.5 & 85.8 & 85.8 \\
MTMD \cite{lin2023multi} & TAFFC'23 & 86.0 & 86.0 & 86.1 & 85.9 \\
MEA \cite{yang2024asynchronous} & TCSVT'24 & 84.4 & 84.6 & 85.2 & 85.1 \\
GeminiFusion \cite{jia2024geminifusion} & ICML'24 & 85.7 & 85.8 & 86.1 & 85.8 \\
EUAR \cite{gao2024enhanced} & MM'24 & 86.3 & 86.3 & \textbf{86.6} & \underline{86.4} \\
DMD \cite{li2023decoupled} & CVPR'24 & 85.8 & 85.8 & 86.0 & 86.1 \\
MDKAT \cite{wang2025mdkat} & TCSVT'25 & 85.6 & 85.6 & \underline{86.5} & \underline{86.4} \\
\midrule
\multicolumn{6}{l}{\textit{Mamba-based Methods}} \\
\midrule
CoupledMamba \cite{li2024coupled} & NeurIPS'24 & - & - & 85.7 & 85.6 \\
Cobra \cite{zhao2025cobra} & AAAI'25 & 82.3 & 82.1 & 81.8 & 81.4 \\
Sigma \cite{wan2025sigma} & WACV'25 & 86.3 & 86.3 & 86.1 & 86.2 \\
\midrule
AlignMamba-1 \cite{li2025alignmamba} & CVPR'25 & \underline{86.9} & \underline{86.9} & \textbf{86.6} & \textbf{86.5} \\
\rowcolor{gray!25} AlignMamba-2 & PR'26 & \textbf{87.0} & \textbf{87.0} & \underline{86.5} & \textbf{86.5} \\
\bottomrule
\end{tabular}
\end{table}

\paragraph{Comparison with Transformer-based Methods}
The first group in the table comprises various Transformer-based fusion models, such as MulT~\cite{tsai2019multimodal}, MISA~\cite{hazarika2020misa}, and Self-MM~\cite{yu2021learning}, which leverage attention mechanisms to capture cross-modal interactions. More advanced methods like MMIM~\cite{han2021improving}, ConFEDE~\cite{yang2023confede}, and MTMD~\cite{lin2023multi} incorporate techniques like mutual information maximization or contrastive learning to enhance fusion quality. Compared to the Transformer-based MoE method~\cite{gao2024enhanced}, our method achieves significant outperformance on the CMU-MOSI dataset and comparable results on the CMU-MOSEI dataset. This demonstrates the superiority of our new architecture, which replaces explicit alignment computation with a more efficient regularization scheme and introduces modality-awareness directly into the fusion core.

\paragraph{Comparison with Mamba-based Methods}
The second group of baselines includes recent methods that have also adopted Mamba for multimodal fusion, such as CoupledMamba~\cite{li2024coupled}, Cobra~\cite{zhao2025cobra}, and Sigma~\cite{wan2025sigma}. These models typically employ strategies like simple concatenation or co-scan mechanisms adapted for Mamba. While they benefit from Mamba's computational efficiency, their performance on these challenging benchmarks is often limited, as seen with Cobra, or does not consistently outperform top-tier Transformer models. For instance, even strong Mamba-based contenders like Sigma achieve results (e.g., 86.3\% Acc on MOSI) that fall short of our proposed method. This performance gap highlights a key insight: simply replacing Transformers with Mamba is insufficient. The architectural limitations of Mamba in handling cross-modal dependencies and multimodal fusion must be explicitly addressed.

\paragraph{Analysis of AlignMamba-2}
AlignMamba-2 demonstrates state-of-the-art performance across both datasets. Notably, on the more challenging MOSEI dataset, it maintains a competitive F1 score of 86.5\%, matching the performance of AlignMamba-1. The consistent high performance can be attributed to our unified framework. The OT and MMD alignment losses ensure that the features fed into the fusion module are already well-aligned from both geometric and statistical perspectives. More importantly, the novel Modality-Aware Mamba layer allows for a more nuanced fusion process by handling modality-specific characteristics and shared patterns. This combination proves to be highly effective for modeling the complex temporal dynamics inherent in multimodal fusion and sentiment analysis tasks.

\subsection{Results on NYUDv2 and MVSA}
\label{sec:static}

To further assess the generalizability of our proposed framework beyond time-series data, we evaluate AlignMamba-2 on two static multimodal classification tasks: scene recognition on NYU-Depth V2 (RGB-D fusion) and sentiment classification on MVSA (Image-Text fusion). The results are summarized in Table~\ref{tab:static_results}. The results demonstrate that AlignMamba-2 consistently outperforms a variety of strong baselines across both datasets. On NYUDv2, it achieves an accuracy of 73.1\%, surpassing recent state-of-the-art methods. Similarly, on the MVSA dataset, it reaches an accuracy of 82.7\%, setting a new benchmark.

\begin{table}[htbp]
\centering
\caption{Results on the NYUDv2 and MVSA datasets.}
\label{tab:static_results}
\begin{tabular}{l|r|cc|cc}
\toprule
\multicolumn{1}{c|}{\multirow{2}{*}{\textbf{Method}}} & \multirow{2}{*}{\textbf{Publication}} & \multicolumn{2}{c|}{\textbf{NYUDv2}} & \multicolumn{2}{c}{\textbf{MVSA}} \\
\multicolumn{1}{c|}{} & & Acc & F1 & Acc & F1 \\
\midrule
\multicolumn{6}{l}{\textit{General Fusion Methods}} \\
\midrule
Late Fusion & - & 69.1 & 68.3 & 76.9 & 75.7 \\
TMC \cite{han2021trusted} & ICLR'21 & 71.1 & 69.8 & 76.1 & 74.6 \\
TokenFusion \cite{wang2022multimodal} & CVPR'22 & 70.8 & 69.5 & 77.4 & 76.0\\
QMF \cite{zhang2023provable} & ICML'23 & 70.1 & 68.7 & 78.1 & 77.2 \\
GeminiFusion \cite{jia2024geminifusion} & ICML'24 & 71.5 & 70.0 & 78.3 & 76.9\\
EAU \cite{gao2024embracing} & CVPR'24 & 72.1 & 70.6 & 79.2 & 78.4 \\
EUAR \cite{gao2024enhanced} & MM'24 & 71.7 & 70.7 & 79.6 & 78.0 \\
MSFN \cite{zhang2025multimodal} & ToMM'25 & - & - & 79.0 & \underline{78.5} \\
\midrule
\multicolumn{6}{l}{\textit{Mamba-based Methods}} \\
\midrule
Cobra \cite{zhao2025cobra} & AAAI'25 & 69.8 & 68.7 & 77.8 & 77.1\\
Sigma \cite{wan2025sigma} & WACV'25 & 71.8 & 70.3 & 78.7 & 78.0\\
\midrule
AlignMamba-1 \cite{li2025alignmamba} & CVPR'25 & \underline{72.4} & \underline{70.9} & \underline{81.1} & 78.4\\
\rowcolor{gray!25} AlignMamba-2 & PR'26 & \textbf{73.1} & \textbf{71.5} & \textbf{82.7} & \textbf{80.2} \\
\bottomrule
\end{tabular}
\end{table}

When comparing with Transformer-based fusion methods such as TokenFusion~\cite{wang2022multimodal} and GeminiFusion~\cite{jia2024geminifusion}, as well as uncertainty-aware models like EAU~\cite{gao2024embracing}, AlignMamba-2 shows a clear advantage. This indicates that our model's effectiveness is not limited to temporal data but extends to tasks requiring the fusion of spatial (image) and semantic (depth, text) information. Compared to the Transformer-based MoE method~\cite{gao2024enhanced}, our method also achieves significant outperformance on both datasets. The performance of Mamba-based counterparts like Cobra~\cite{zhao2025cobra} and Sigma~\cite{wan2025sigma}, while superior to a simple Late Fusion baseline, does not reach the levels of top-performing methods. This again reinforces our core argument: a naive application of Mamba is insufficient for complex fusion tasks. Our previous work, AlignMamba-1, achieves competitive results, but is surpassed by AlignMamba-2, highlighting the benefits of our refined alignment strategy and the novel modality-aware architecture.

In essence, the success on these static tasks validates that our dual alignment strategy and the Modality-Aware Mamba layer are general mechanisms for bridging the heterogeneity gap, regardless of whether the data has a temporal structure. The model effectively learns to align and fuse different data structures (patch-level features from images and depth maps, and token-level features from text), showcasing its robustness and wide applicability.

\subsection{Computational Efficiency Analysis}
\label{sec:efficiency}
To evaluate the computational efficiency of our proposed method, we conduct a comparative analysis of GPU memory consumption and inference latency. We benchmark AlignMamba-2 against our prior work, AlignMamba-1~\cite{li2025alignmamba}, and a representative Transformer-based model, MulT~\cite{tsai2019multimodal}. To ensure a fair comparison, our analysis focuses exclusively on the fusion module, excluding the unimodal encoders. For all experiments, the batch size is set to 1, and we vary the input sequence length from 10k up to 100k. Each data point is the average of 10 independent runs to ensure stable and reliable measurements.

\begin{figure}[htbp]
\centering
\includegraphics[width=0.7\linewidth]{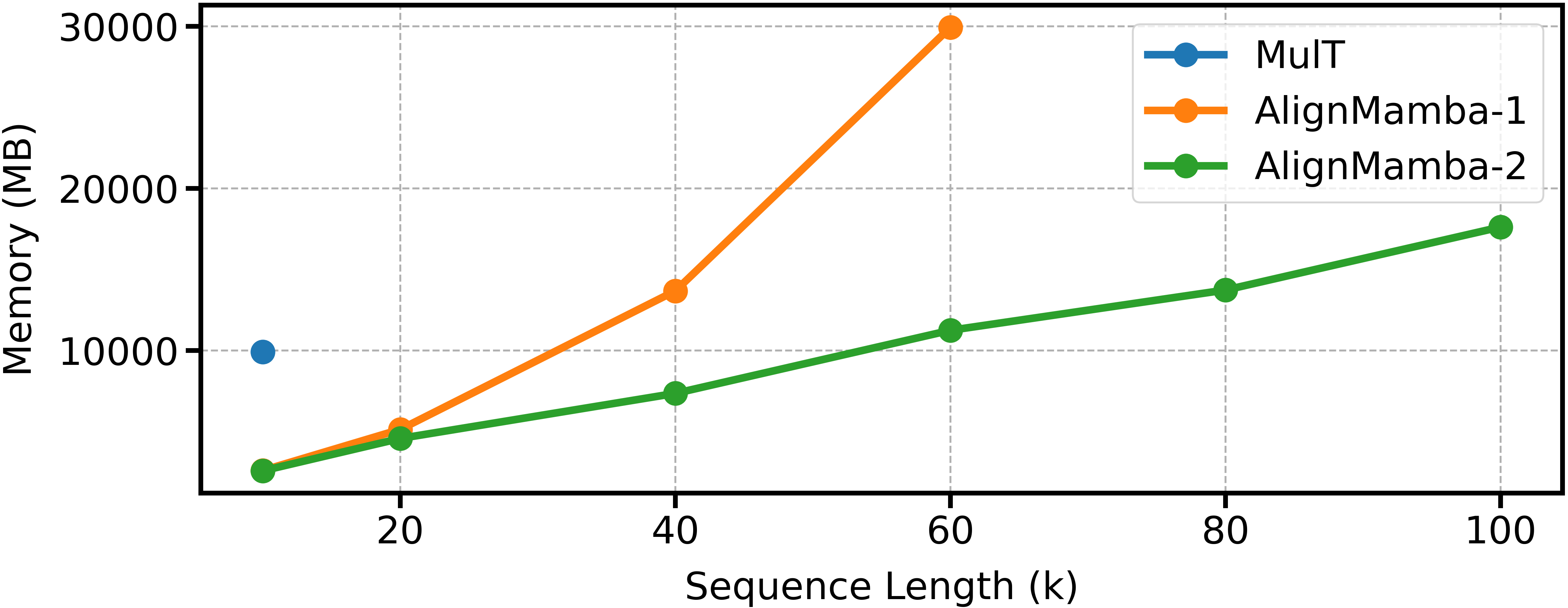}
\caption{GPU memory consumption (in MB) as a function of increasing sequence length. AlignMamba-2 demonstrates superior scalability compared to both Transformer-based and explicit-alignment-based Mamba models.}
\label{fig:memory}
\end{figure}

\paragraph{GPU Memory Usage}
Figure~\ref{fig:memory} illustrates the GPU memory usage of the three models. The Transformer-based MulT consumes approximately 10 GB of memory for a 10k sequence and encounters an out-of-memory error at 20k. This behavior starkly highlights the quadratic complexity of the self-attention mechanism, which leads to a dramatic increase in memory requirements as the sequence length grows.

Both AlignMamba-1 and AlignMamba-2 exhibit comparable memory consumption at shorter sequence lengths (10k and 20k). However, as the length extends to 40k and 60k, the memory footprint of AlignMamba-1 increases significantly, culminating in an OOM error around 80k. This is because AlignMamba-1 explicitly computes the Optimal Transport (OT) matrix during inference, a process that incurs substantial resource costs for long sequences. In contrast, AlignMamba-2 leverages OT distance as a regularization loss \textit{during training} and does not require this computation at inference time. Consequently, it maintains a linear and highly efficient memory profile, successfully processing sequences up to 100k and beyond.

\begin{figure}[htbp]
\centering
\includegraphics[width=0.7\linewidth]{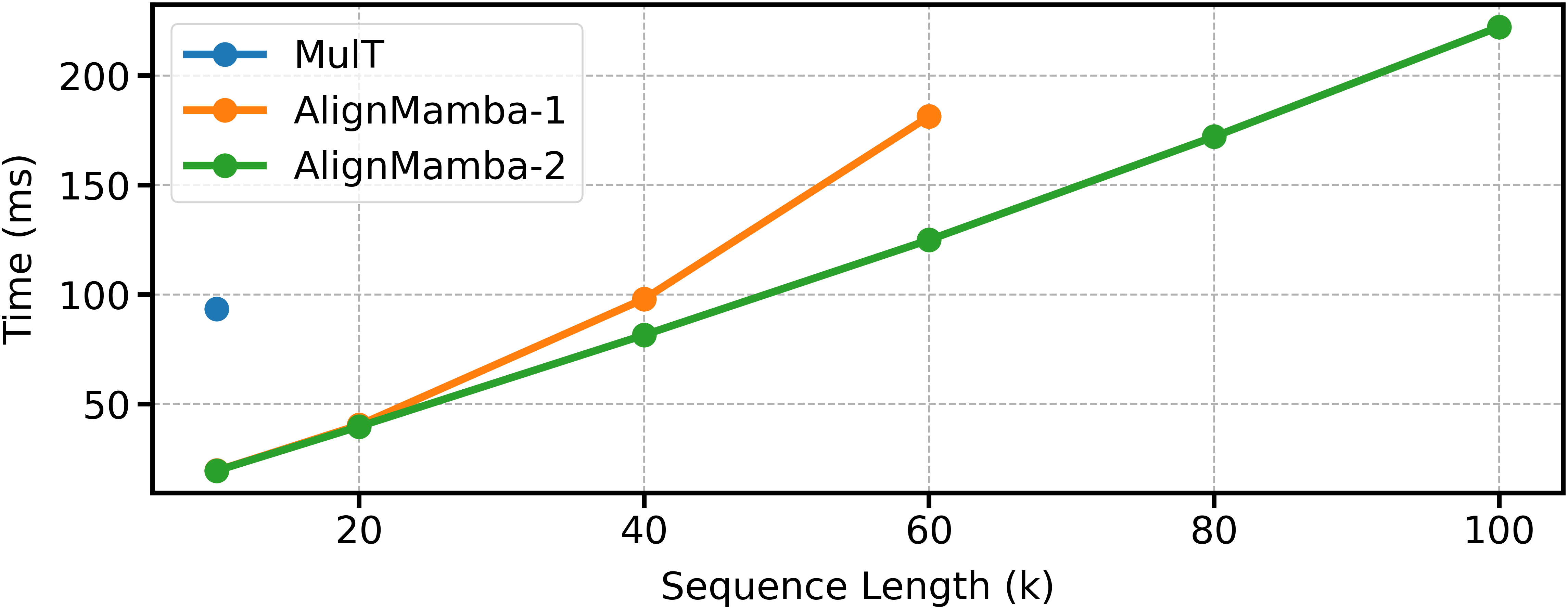}
\caption{Inference latency (in milliseconds) as a function of increasing sequence length. AlignMamba-2 maintains a linear increase in inference time, showcasing its efficiency for long-sequence processing.}
\label{fig:time}
\end{figure}

\paragraph{Inference Time}
Figure~\ref{fig:time} presents the comparison of inference latency. MulT requires nearly 100 milliseconds to process a 10k sequence, underscoring the severe latency bottleneck of Transformer-based fusion. AlignMamba-1 offers a substantial improvement, processing a 40k sequence in a similar amount of time. AlignMamba-2, however, demonstrates even greater efficiency. It not only handles much longer sequences but also maintains a near-linear growth in processing time as the sequence length increases. This result shows AlignMamba-2 achieves remarkable gains in both speed and scalability, making it a highly practical solution for real-world, long-sequence multimodal applications.

\subsection{Ablation Studies}
\label{sec:ablation}
To thoroughly investigate the individual contributions of the core components in AlignMamba-2, we conduct a series of ablation studies across all four datasets. We systematically remove or modify key parts of our model: the dual alignment loss, the Mixture-of-Experts mechanism in the fusion layers, and the deterministic routing strategy. The results are detailed in Table~\ref{tab:ablation}.

\begin{table}[htbp]
\centering
\caption{Ablation study of AlignMamba-2 on all four datasets. We report the performance (Accuracy and F1-Score) of the full model and its variants. "w/o" stands for "without", and "w/" stands for "with". Best results are in bold.}
\label{tab:ablation}
\begin{tabular}{l|cc|cc|cc|cc}
\toprule
 & \multicolumn{2}{c}{\textbf{MOSI}} & \multicolumn{2}{c}{\textbf{MOSEI}} & \multicolumn{2}{c}{\textbf{NYUDv2}} & \multicolumn{2}{c}{\textbf{MVSA}} \\
\cmidrule(lr){2-3} \cmidrule(lr){4-5} \cmidrule(lr){6-7} \cmidrule(lr){8-9}
 & Acc & F1 & Acc & F1 & Acc & F1 & Acc & F1 \\
\midrule
\rowcolor{gray!25} AlignMamba-2 & \textbf{87.0} & \textbf{87.0} & \textbf{86.5} & \textbf{86.5} & \textbf{73.1} & \textbf{71.5} & \textbf{82.7} & \textbf{80.2} \\
\midrule
w/o dual alignment loss & 84.5 & 84.4 & 84.1 & 84.0 & 71.0 & 69.5 & 79.2 & 78.1 \\
w/o Mixture-of-Experts & 85.8 & 85.7 & 85.3 & 85.2 & 72.0 & 70.4 & 80.5 & 78.2 \\
w/o both & 82.3 & 82.1 & 81.8 & 81.4 & 69.8 & 68.7 & 77.8 & 77.1 \\
\midrule
w/ Learnable Routing & 86.6 & 86.4 & 86.1 & 86.2 & 72.8 & 71.2 & 82.2 & 79.5 \\
\bottomrule
\end{tabular}
\end{table}

\paragraph{Impact of the Dual Alignment Loss}
The first variant, "w/o dual alignment loss," removes both the OT and MMD regularization terms, meaning the unimodal features are directly concatenated and fed into the fusion module without any explicit alignment constraints. As shown in the table, this leads to the most significant performance degradation across all benchmarks. For instance, on CMU-MOSI, the F1 score drops by 2.6\%, and on NYUDv2, it drops by 2.0\%. This substantial decline underscores the critical importance of pre-fusion alignment. Without it, the Mamba-based fusion module struggles to bridge the large semantic and statistical gaps between heterogeneous modalities, validating our initial hypothesis that alignment is a prerequisite for effective fusion, especially within a sequential modeling framework like Mamba.

\paragraph{Impact of the Mixture-of-Experts Mechanism}
In the "w/o Mixture-of-Experts" setting, we replace our proposed Modality-Aware Mamba layers with standard, vanilla Mamba layers. In this configuration, the model still benefits from the dual alignment loss but loses the ability to process tokens in a modality-specific manner during fusion. The results show a consistent, smaller performance drop compared to removing the alignment loss. For example, the F1 score decreases by 1.3\% on MOSI and 1.1\% on NYUDv2. This demonstrates that while proper alignment is crucial, enabling the fusion backbone to explicitly model both modality-specific and modality-invariant information through the MoE architecture provides a significant additional benefit, leading to more nuanced and powerful representations. Removing both components ("w/o both"), which effectively reduces our model to a simple Mamba baseline, results in a drastic performance collapse, confirming that both alignment and awareness are indispensable.

\paragraph{Impact of Deterministic vs. Learnable Routing}
Finally, we explore the design of our MoE routing mechanism. The "Learnable Routing" variant replaces our deterministic, modality-based routing with a conventional learnable gating layer that dynamically decides which expert(s) to activate for each token. The results show that the performance on all datasets is slightly reduced, but still higher than that without the MoE architecture. This suggests that for multimodal fusion, the modality of a token is a powerful and reliable signal for routing. A learnable gate introduces additional complexity and potential optimization challenges, without providing a clear benefit over our simpler, more direct deterministic approach. For learnable routing without additional constraints, it is challenging to perform modality-specific modeling, which leads to multimodal representations that are insufficiently discriminative. This finding validates our design choice to leverage prior knowledge of token modality for efficient and effective expert selection.

\subsection{Cross-Dataset Generalization Analysis}
\label{sec:generalization}
A robust multimodal fusion model should not only perform well within a specific data distribution but also demonstrate strong generalization capabilities to unseen, out-of-domain data. To evaluate this, we conduct a cross-dataset generalization experiment. Due to inconsistencies in the official visual feature dimensions between the CMU-MOSI and CMU-MOSEI datasets~\cite{zadeh2016mosi,zadeh2018multimodal}, we focus this analysis on a bimodal (audio-text) setup to ensure a fair and direct comparison. Specifically, we train our model and two representative baselines (Cobra~\cite{zhao2025cobra} and GeminiFusion~\cite{jia2024geminifusion}) on one dataset and test their performance on both the original (in-domain) and the other (out-of-domain) dataset in a zero-shot manner. The results of this rigorous evaluation are presented in Table~\ref{tab:generalization}.

\begin{table}[htbp]
\centering
\caption{Cross-dataset generalization performance on a bimodal (audio-text) setup. The table shows the in-domain and out-of-domain (zero-shot) performance for models trained on MOSI and MOSEI, respectively. Best results are in bold.}
\label{tab:generalization}
\begin{tabular}{l|cc|cc|cc|cc}
\toprule
Training dataset & \multicolumn{4}{c|}{MOSI} & \multicolumn{4}{c}{MOSEI} \\
\cmidrule(lr){1-2} \cmidrule(lr){2-5} \cmidrule(lr){6-9}
Test dataset & \multicolumn{2}{c|}{MOSI} & \multicolumn{2}{c|}{MOSEI} & \multicolumn{2}{c|}{MOSEI} & \multicolumn{2}{c}{MOSI} \\
\cmidrule(lr){1-2} \cmidrule(lr){2-3} \cmidrule(lr){4-5} \cmidrule(lr){6-7} \cmidrule(lr){8-9}
Model & Acc & F1 & Acc & F1 & Acc & F1 & Acc & F1 \\
\midrule
Cobra & 80.2 & 80.0 & 70.5 & 70.3 & 79.5 & 79.2 & 70.1 & 70.0 \\
GeminiFusion & 83.2 & 83.1 & 74.6 & 74.6 & 82.6 & 82.5 & 75.0 & 75.1 \\
AlignMamba-1 & 83.8 & 83.8 & 75.6 & 76.8 & 83.2 & 83.1 & 76.4 & 76.3 \\
\rowcolor{gray!25} AlignMamba-2 & \textbf{84.0} & \textbf{83.9} & \textbf{77.4} & \textbf{77.5} & \textbf{83.4} & \textbf{83.2} & \textbf{77.0} & \textbf{77.1} \\
\bottomrule
\end{tabular}
\end{table}

As expected, all models experience a significant performance drop when tested on an out-of-domain dataset, which is attributable to the inherent domain shift in speaker characteristics, vocabulary, and emotional expression styles between MOSI and MOSEI. However, the key finding is the \textit{relative} robustness of AlignMamba-2.

When trained on MOSI and tested on MOSEI, AlignMamba-2 achieves an F1 score of 77.5\%, substantially outperforming GeminiFusion (74.6\%) and Cobra (70.3\%). A similar trend is observed in the reverse scenario (trained on MOSEI, tested on MOSI), where AlignMamba-2 again demonstrates superior performance. This indicates that our model learns more transferable and generalizable representations compared to the baselines.

Furthermore, while AlignMamba-2 achieves comparable in-domain performance to AlignMamba-1, it delivers superior results in out-of-domain settings, demonstrating enhanced cross-dataset generalization and robustness. We attribute this improvement to two key factors: First, unlike the explicit OT transport matrix computation in AlignMamba-1, the implicit OT loss regularization in AlignMamba-2 encourages the model to learn more robust, intrinsic representations. Second, the Modality-Aware MoE facilitates the disentanglement of modality-specific features, leading to improved transferability.

\subsection{Hyperparameter Analysis}
\label{sec:hyperparams}
\begin{figure}[htbp]
\centering
\includegraphics[width=0.9\textwidth]{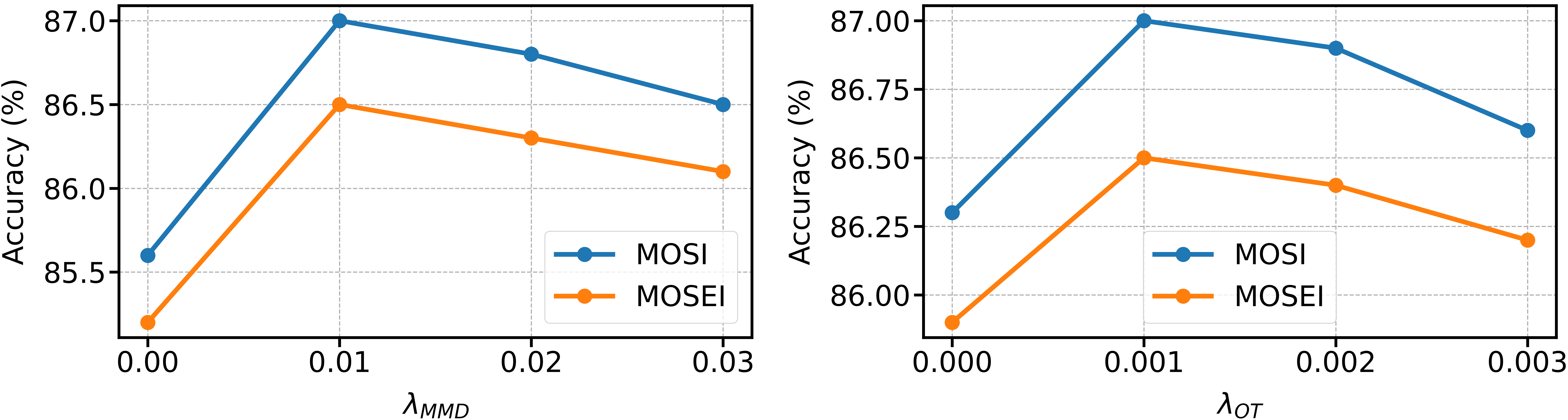}
\caption{Sensitivity analysis of model accuracy with respect to the MMD loss weight ($\lambda_{MMD}$) and the OT loss weight ($\lambda_{OT}$) on the MOSI and MOSEI datasets.}
\label{fig:hyperparams}
\end{figure}

In this section, we investigate the impact of the key hyperparameters that govern our dual alignment strategy: the MMD loss weight, $\lambda_{\text{MMD}}$, and the OT loss weight, $\lambda_{\text{OT}}$. These weights control the strength of the alignment regularization relative to the main task objective. Figure~\ref{fig:hyperparams} illustrates the model's performance on the CMU-MOSI and CMU-MOSEI datasets as these two parameters are varied.

The left panel of the figure shows the sensitivity to $\lambda_{\text{MMD}}$ while keeping $\lambda_{\text{OT}}$ at its optimal value. When $\lambda_{\text{MMD}}$ is set to zero, the model's performance is sub-optimal, clearly indicating that the global statistical alignment provided by the MMD loss is beneficial. As the weight increases from zero, performance improves significantly. The model achieves its peak performance on both datasets at $\lambda_{\text{MMD}} = 0.01$, reaching a top accuracy of 87.0\% on MOSI. Further increasing the weight beyond this point leads to a slight but consistent decline in performance. This trend suggests that while MMD alignment is crucial, overly strong regularization can start to interfere with the primary task objective.

A similar pattern is observed for the OT loss weight, $\lambda_{\text{OT}}$, as shown in the right panel, with $\lambda_{\text{MMD}}$ held at its optimal setting. Performance without the OT loss is notably lower than the peak, confirming the positive contribution of the local geometric alignment. The model reaches the same optimal performance at a value of $\lambda_{\text{OT}} = 0.001$. As the weight increases past this point, we again observe a gentle degradation in accuracy, reinforcing the idea that balanced regularization is key to achieving the best results.

Overall, these results demonstrate the importance of both alignment losses for guiding the model toward better fusion. They also show that while the model is sensitive to these hyperparameters, it is robust within a reasonable range around their optimal values. The fact that the optimal weights are relatively small suggests that the alignment objectives act as effective regularizers that steer the learning process without dominating the task-specific loss.

\subsection{Case Study}
\begin{figure*}[htbp]
\centering
\includegraphics[width=\linewidth]{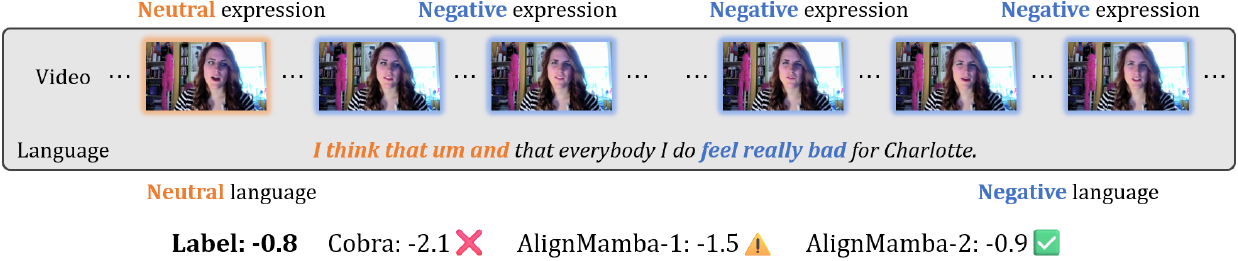}
\caption{Qualitative analysis of a sample from the CMU-MOSI dataset.}
\label{fig:case}
\end{figure*}

To provide a more intuitive understanding of how AlignMamba-2 improves multimodal fusion, we conduct a qualitative case study on a representative sample from the CMU-MOSI dataset. As illustrated in Figure~\ref{fig:case}, we visualize the temporal evolution of sentiment in the video and text modalities, and compare the sentiment predictions of our model against two baselines: Cobra~\cite{zhao2025cobra} and our prior work, AlignMamba-1~\cite{li2025alignmamba}.

The selected sample presents a classic case of inter-modal temporal misalignment. The visual modality (e.g., facial expressions) exhibits a neutral sentiment at the beginning before shifting to a sustained negative emotion for the remainder of the clip. In contrast, the textual modality remains neutral for the first half of the utterance and only introduces explicitly negative words towards the end. The ground-truth sentiment label for the entire sample is -0.8, indicating a weakly negative to neutral sentiment.

Cobra predicts a sentiment score of -2.1, indicating a strong negative emotion. Lacking an effective cross-modal alignment mechanism, it appears to over-emphasize the strong negative cues present at the end of each modality, failing to temper this with the initial neutral context. This results in a prediction that significantly deviates from the ground truth. AlignMamba-1, which incorporates an explicit OT-based alignment module, yields a more accurate prediction of -1.5. The alignment mechanism enables it to better correlate the temporal segments across modalities, leading to a more holistic understanding than Cobra. However, its prediction still leans more negative than the ground truth. In stark contrast, our proposed AlignMamba-2 delivers the most accurate prediction of -0.9. This superior result can be attributed to its two-fold architectural advancements. First, the dual alignment strategy (OT and MMD) ensures a more comprehensive alignment of the underlying feature distributions, allowing the model to correctly balance the influence of both the neutral and negative segments. Second, the Modality-Aware Mamba layer, with its specialized experts, can more effectively process the distinct temporal dynamics of visual and textual sentiment expression. By integrating these aligned and modality-aware insights, AlignMamba-2 achieves a nuanced fusion that closely mirrors the subtle, overall sentiment of the sample. This case study vividly demonstrates the practical benefits of our model's design in handling complex, real-world multimodal interactions.
\section{Conclusion}
In this paper, we introduced AlignMamba-2, a novel framework for multimodal fusion and sentiment analysis. First, we proposed an efficient dual alignment strategy, using Optimal Transport distance and Maximum Mean Discrepancy as regularization to prompt comprehensive pre-fusion alignment without any inference cost. Second, we developed a novel Modality-Aware Mamba layer that leverages a Mixture-of-Experts design to explicitly model both modality-specific characteristics and shared cross-modal patterns. Through extensive experiments on a diverse set of dynamic and static multimodal benchmarks, we demonstrated that AlignMamba-2 consistently achieves state-of-the-art performance. Furthermore, our detailed efficiency analysis confirmed its significant advantages in memory consumption and inference speed for long-sequence tasks.

However, consistent with the "no free lunch" principle, the robust alignment capability comes with a trade-off. Specifically, calculating OT and MMD losses during the training phase introduces additional computational overhead compared to simple baseline methods. Our work not only presents a powerful solution for multimodal learning and sentiment analysis but also highlights a critical direction for future research: exploring linear-complexity alignment metrics or sparse mechanisms to achieve unified efficiency across both training and inference phases, ensuring discriminative multimodal representations with minimal computational cost.

\section{Acknowledgement}
This paper is supported by the China Postdoctoral Science Foundation (Grant No. 2025M781481) and the National Natural Science Foundation of China (Grant No. 62236006).

\bibliography{main}

\end{document}